%% file: main.tex
\pdfoutput=1
\documentclass{INTERSPEECH2023}

\interspeechcameraready

\title{MeetEval: A Toolkit for Computation of Word Error Rates for Meeting Transcription Systems}

\name{Thilo von Neumann$^1$, Christoph Boeddeker$^1$, Marc Delcroix$^2$, Reinhold Haeb-Umbach$^1$}
\address{
  $^1$Paderborn University, Germany\\
  $^2$NTT Corporation, Japan}
\email{vonneumann@nt.upb.de, boeddeker@nt.upb.de, marc.delcroix@ieee.org, haeb@nt.upb.de}

\input{macros.tex}
\input{tikz_style.tex}
\input{symbols.tex}
\input{glossaries}

\begin{document}

\maketitle
 
\begin{abstract}

MeetEval is an open-source toolkit to evaluate  all kinds of meeting transcription systems.
It provides a unified interface for the computation of commonly used Word Error Rates (WERs), specifically cpWER, ORC-WER and MIMO-WER along other WER definitions.
We extend the cpWER computation by a temporal constraint to ensure that only words are identified as correct when the temporal alignment is plausible.
This leads to a better quality of the matching of the hypothesis string to the reference string that more closely resembles the actual transcription quality, and a system is penalized if it provides poor time annotations.
Since word-level timing information is often not available, we present a way to approximate exact word-level timings from segment-level  timings (e.g., a sentence) and show that the approximation leads to a similar WER as a matching with exact word-level annotations.
At the same time, the time constraint leads to a speedup of the matching algorithm, which outweighs the additional overhead caused by processing the time stamps.

\end{abstract}
\section{Introduction}\label{introduction}

\vphantom{\gls{MIMO-WER}}The \gls{WER} is a common metric for evaluation of \gls{ASR} systems.
While the standard \gls{WER}, defined as the number of wrongly recognized words divided by the total number of words, is used for single-speaker \gls{ASR}, it is not directly applicable to multi-speaker meeting transcription.
Modifications of the standard \gls{WER}, such as the \gls{cpWER} and \gls{ORC-WER}, have been proposed in order to assess meeting transcription performance with a \gls{WER} metric.

There is, however, no toolkit that allows easy computation of these \gls{WER} measures.
The NIST Scoring Toolkit (SCTK)\footnote{\url{https://github.com/usnistgov/SCTK}} implements some metrics.
Those tools are, however, not built for modern meeting transcription systems, suffer from large memory usage and lack support for some common system output formats.
The \gls{cpWER} is available in the Kaldi speech recognition toolkit \cite{Povey2011_KaldiSpeechRecognition}, but not easily accessible.
WER metrics that emerged recently, such as the \gls{ORC-WER} \cite{Sklyar2022_MultiTurnRNNTStreaming} or MIMO-WER \cite{vonNeumann2023_WordErrorRate}, have no published implementation outside of MeetEval\footnote{\url{https://github.com/fgnt/meeteval}}.

We present MeetEval with the aim of providing a unified and easy-to-use interface for different WER definitions, including widely used metrics (\gls{cpWER}) and newly emerging metrics (\gls{ORC-WER}).
By this we hope to facilitate assessing the performance of such systems in a reproducible and comparable way.
Having different metrics implemented in the same toolkit with the same interface allows easy switching between metrics and thus a deeper analysis of the \gls{ASR} errors.

We additionally propose the \gls{tcpWER}, an extension of the \gls{cpWER} that identifies words as correct or substituted only when the temporal alignment with the reference transcription is plausible.
While a similar idea has been mentioned and used in the past \cite{Fiscus2006_MultipleDimensionLevenshtein,Morris2004_WERRILMER}, its impact on the metric has never been assessed for meeting transcription.
Since for modern systems, precise word level temporal annotations are often not available (as opposed to what was assumed in earlier works \cite{Morris2004_WERRILMER}), we propose a way to approximate the word-level annotations from coarse segment-level annotations using a word-length-based heuristic.
We analyse and discuss practical issues for reference annotations and show that a collar is necessary to obtain a \gls{WER} that matches the \gls{WER} one would get with high precision annotations.
Furthermore, the time constraint leads to a speedup of the matching algorithm.

\section{Scenario}\label{sec:scenario}

\begin{figure*}[h]
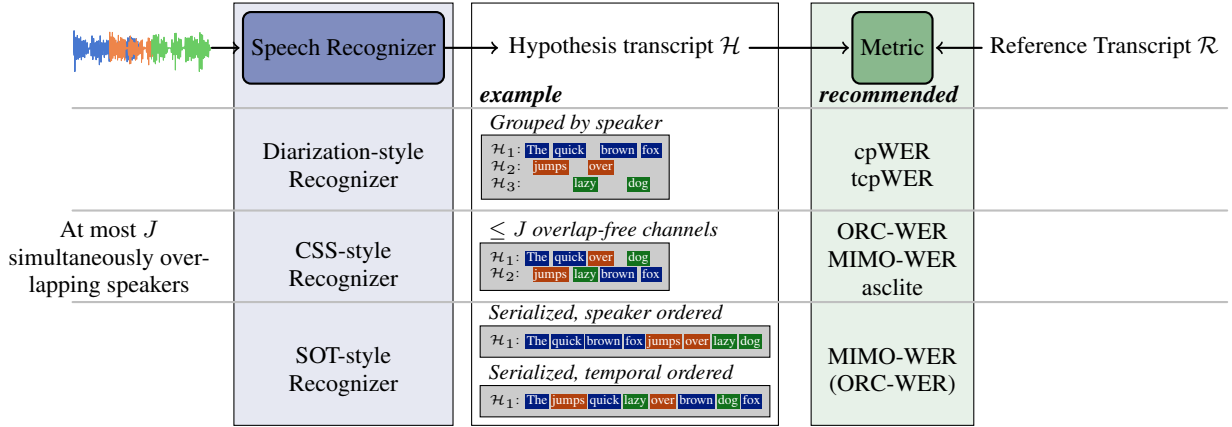

    \centering
    \include{tikz/scenario}
    \vspace{-2em}
    \caption{
        MeetEval is tailored to the meeting transcription scenario.
        The input to a speech recognizer contains speech of multiple speakers with potential overlap.
        The output formats of the different recognizer styles are visualized using a simple example. 
        Letters represent words, words represent segments/utterances and colors represent speakers.
    }\label{fig:scenario}
    \vspace{-2em}
\end{figure*}

MeetEval is tailored to the evaluation of systems with \gls{WER} that transcribe meetings.
\cref{fig:scenario} gives an overview of the most commonly used recognizers, their output formats and the appropriate metrics.
A typical recognizer receives a recording of a meeting and outputs a transcript, where different systems produce different representations and levels of detail.
The hypothesis transcript is then evaluated against a reference transcript with a (\gls{WER}-based) metric.

The recording at the input is typically relatively long (at least a few minutes and up to hours) and contains multiple speakers with complex unpredictable speaking patterns which may lead to speech overlaps and arbitrarily long speech pauses.
The hypothesis transcript consists of one or multiple streams $\thyp_1,\dots,\thyp_\nchn$, where each stream contains a part of the transcript.
Words can be grouped arbitrarily on these streams, but they are commonly grouped either
\begin{itemize}
    \item in  \emph{Diarization-style}, i.e., \emph{by speaker}, as in \cite{Watanabe2020_CHiME6ChallengeTackling},
    \item in \emph{CSS-style}, i.e., such that overlapping speech is placed on different streams inspired by the Continuous Speech Separation (CSS) pipeline \cite{Chen2020_ContinuousSpeechSeparation}, or
    \item in \emph{SOT-style}, where all transcripts are serialized into one output stream as it is done by systems trained with Serialized Output Training (SOT) \cite{Kanda2020_SerializedOutputTraining,Kanda2022_StreamingMultiTalkerASR}.
    The stream is often serialized in temporal order, but the recognizer can in principle choose any order, e.g., output several sentence from one speaker before moving back in time to transcribe another speaker.
\end{itemize}
Diarization- and CSS-stlye recognizers typically deliver segment-level time annotations, where a segment is a group of words that are close to each other (e.g., an utterance).
The segment boundaries are found by the recognizer.

We assume that the references contain speaker labels, and for some metrics we also assume a known word order within a segment and segment-level timings of reasonable accuracy from which we approximate word-level timings.

\section{Word Error Rates for Meeting Transcription}\label{sec:word-error-rates-for-meeting-transcription}

MeetEval implements a variety of word error rate metrics for the evaluation of meeting transcription systems.
This section gives a short overview of the core metrics,
%
%
where we use the following notation: 
We use $\tref_{\ispk,\iutt}$ to denote the $\iutt$-th reference utterance of speaker $\ispk$ and $\thyp_{\ichn,\iutt'}$ to denote the $\iutt'$-th segment of the $\ichn$-th system/hypothesis output stream.
$\nutt_{\tref_{\ispk}}$ is the number of utterances/segments in $\tref_{\ispk}$.
The references/segments are ordered by begin time.
Subscripts are neglected where they are not important or applicable.
For any word sequence $\tref$, $|\tref|$ is the number of words and $\tref(\iref)$ is the $\iref$-th word in $\tref$.
We denote sequence concatenation with $\oplus$.

Given the scenario from \cref{sec:scenario}, an assignment problem arises because the mapping of utterances to outputs is not unique, e.g., the streams of a Diarization-style recognizer can be permuted along the speaker axis.
This assignment problem is solved in different ways by different \gls{WER} definitions.

\subsection{Standard WER}\label{sec:standard-wer}

The standard \gls{WER} is a common metric to evaluate the performance of conventional single-speaker speech recognizers.
It can only be applied to pairs of a single reference utterance $\tref$ and a single hypothesis segment $\thyp$.
It is defined as the number of word errors in relation to the number of total words in $\tref$:
\begin{align}
    \text{WER} = \frac{\text{\#total word errors}}{\text{\#total words}} = \frac{\sum_{\tref, \thyp} \lev(\tref, \thyp)}{\sum_{\tref}|\tref|},   \label{eq:stdwer}
\end{align}
where $\lev(\tref,\thyp)$ is the Levenshtein distance \cite{Levenshtein1965_BinaryCodesCapable} between $\tref$ and $\thyp$,  and $\sum_{\tref, \thyp}$ is the summation of all pairs of reference and hypothesis in a dataset.
The Levenshtein distance $\lev$ is the minimum number of word edit operations (substitution, deletion, insertion) to change the hypothesis sequence into the reference sequence. 
Note, though, that the decomposition into substitutions, insertions and deletions is not unique.

The Wagner-Fischer algorithm \cite{Wagner1974_StringtoStringCorrectionProblem} can be used to efficiently compute $\lev$.
It creates a two-dimensional distance matrix $\lm\in\reals^{(|\tref|+1)\times(|\thyp|+1)}$, where each entry $\lm_{\iref,\ihyp} ~\forall~ \iref \in \{0, \dots, |\tref|\}, \ihyp \in \{0, \dots, |\thyp|\}$ is the Levenshtein distance of the sub-sequences up to the $\iref$-th and $\ihyp$-th word of $\tref$ and $\thyp$, respectively, where $r=0$ or $h=0$ means comparing with the empty string.
The computation starts with $\lm_{0, h}=h$ and $\lm_{r, 0}=r$, and the entries of $\lm_{\iref,\ihyp}$ are recursively computed:
\begin{align}
    \lm_{\iref, \ihyp} &= \min
    \begin{cases}
    \lm_{\iref-1, \ihyp - 1} + \costcorr & \text{if } \tref(\iref)=\thyp(\ihyp) \\
    \lm_{\iref-1, \ihyp - 1}  + \costsub & \text{if } \tref(\iref)\neq\thyp(\ihyp) \\
    \lm_{\iref, \ihyp - 1} + \costins\\
    \lm_{\iref-1, \ihyp} + \costdel,
    \end{cases} \label{eq:lev}
\end{align} 
where $\costcorr$, $\costsub$, $\costins$ and $\costdel$ are the costs of a correct match, a substitution, an insertion, and a deletion operation, respectively.
The costs are typically chosen as $\costsub=\costins=\costdel=1$ and $\costcorr=0$.
The final value is $\lev(\tref,\thyp)=\lm_{|\tref|,|\thyp|}$.

The standard \gls{WER} is the basis for all other \gls{WER} definitions.
In the following, we only state the Levenshtein distance for simplicity.
The corresponding \gls{WER} is the Levenshtein distance divided by the total number of words, see \cref{eq:stdwer}.

\subsection{cpWER}\label{sec:cpwer}

One common metric for evaluation of \emph{Diarization-style} output is the \gls{cpWER} \cite{Watanabe2020_CHiME6ChallengeTackling}.
The reference streams $\tref_1,...,\tref_\nspk$ and hypothesis streams $\thyp_1,...,\thyp_\nspk$ are grouped by $\nspk$ speakers, where segment/utterance transcriptions are concatenated.
Then, the standard \gls{WER} is computed between the hypothesis and all permutations $\pi\in P(\nspk)$ of reference streams.
The permutation which minimizes the \gls{WER} is reported:
\begin{align}
    \lev^{\text{(cp)}}=\min_{\pi\in P(\nspk)} \sum_{\ispk=1}^\nspk \mathrm{lev}\left(\bigoplus_{\iutt=1}^{\nutt_{\tref_{\pi(\ispk)}}} \tref_{\pi(\ispk), \iutt}, \bigoplus_{\iutt'=1}^{\nutt_{\thyp_\ispk}} \thyp_{\ispk,\iutt'}\right).
    \label{eq:cp}
\end{align}
Empty streams are inserted for the reference or hypothesis when the number of speakers is over- or under-estimated, respectively.

The naive computational complexity is $\bigO(\nspk!)$ since $|P(\nspk)|=\nspk!$.
MeetEval, however, uses the Hungarian algorithm \cite{Kuhn1955_HungarianMethodAssignment,Munkres1957_AlgorithmsAssignmentTransportation} to find the permutation in polynomial time.
The computation of all $\nspk^2$ pairwise Levenshtein distances has a complexity of $\bigO(\nspk^2)$ and dominates the total run time in practice.

\subsection{MIMO-WER}\label{sec:mimo-wer}

The recently proposed MIMO-WER \cite{vonNeumann2023_WordErrorRate} measures the \gls{WER} for multi-speaker scenarios without considering segment-level speaker assignment errors.
It solves the assignment problem on a segment level, according to
\begin{align}
    \lev^{\text{(MIMO)}}=\min_{\rho_1, \dots, \rho_J} \sum_{\ichn=1}^\nchn \mathrm{lev}\left(\bigoplus_{(\ispk,\iutt) \in \rho_\ichn} \mkern-12mu\tref_{\ispk,\iutt}, \bigoplus_{\iutt'=1}^{\nutt_{\thyp,\ichn}}\thyp_{\ichn,\iutt'}\right),\label{eq:mimo}
\end{align}
where $\rho_\ichn$ contains index pairs (speaker index $\ispk$ and utterance index $\iutt$) of the reference utterances that are assigned to the $\ichn$-th hypothesis.
Each reference is assigned exactly once and the references must be sorted by begin time within a speaker.
The order of utterances is unconstrained across speakers, which is necessary when evaluating \emph{SOT-style} systems that can output utterance of different speakers in an arbitrary order.
%
The MIMO-WER penalizes splitting utterances over multiple streams since a reference utterance is always assigned to one output stream continuously, 
The distance $\lev^{\text{(MIMO)}}$ is computed with a multi-dimensional variant of the Levenshtein distance dynamic programming algorithm \cite{vonNeumann2023_WordErrorRate}.

\subsection{Optimal Reference Combination WER (ORC-WER)}
\label{sec:optimal-reference-combination-wer-orc-wer}

The \gls{ORC-WER} \cite{Sklyar2022_MultiTurnRNNTStreaming,Raj2022_ContinuousStreamingMultiTalker} is a special case of the MIMO-WER which keeps the temporal order across speakers intact.
It can be computed with \cref{eq:mimo} by ignoring the reference speaker labels:
\begin{align}
    \mathrm{lev}^{\text{(ORC)}} = \min_{\rho_1, \dots, \rho_J} \sum_{\ichn=1}^\nchn \mathrm{lev}\left(\bigoplus_{\iutt\in\rho_\ichn}\tref_{\iutt},  \bigoplus_{\iutt'=1}^{\nutt_{\thyp,\ichn}}\thyp_{\ichn,\iutt'}\right).
\end{align}
%
The complexity does not depend on the number of reference speakers and is polynomial in the number of utterances \cite{vonNeumann2023_WordErrorRate}.

Note that ORC-WER and MIMO-WER only differ in certain edge cases when the system modifies the order of utterances (e.g., certain SOT systems \cite{Kanda2020_SerializedOutputTraining}) or where the utterance order is ambiguous in the reference.
The ORC-WER overestimates the \gls{WER} in these cases \cite{vonNeumann2023_WordErrorRate}.
When the utterance order is well defined, the ORC-WER is preferred because its complexity is significantly smaller when the number of speakers $\nspk$ is large and the number of system output streams $\nchn$ is small.

\subsection{When to use which WER?}

\cref{fig:scenario} gives an overview of widely used recognizer styles and which \gls{WER} we recommend for each.
For \emph{Diarizaiton-style} recognizers, the \gls{cpWER} can be computed and gives a good indication of the system performance.
The \gls{ORC-WER} and \gls{MIMO-WER} can theoretically be computed to gain insight what the \gls{WER} would be if all speaker labels were correct.
But in practice, this is often infeasible because the complexity explodes with increasing number of system output streams $\nchn$.

For \emph{CSS-style} and \emph{SOT-style} recognizers, where the output is not grouped by speakers, \gls{ORC-WER} or \gls{MIMO-WER} have to be used.
While \gls{MIMO-WER} is desired in such a situation, this is often infeasible (large $\nchn$).
\gls{ORC-WER} is well suited and often equal to the \gls{MIMO-WER} when the output follows the temporal order of the physical signal and has a reasonable execution time when the number of system output streams $\nchn$ is small.

An \emph{SOT-style} recognizer is one of the few exceptions where the recognizer can jump backward in time so that the \gls{MIMO-WER} is the only applicable metric.
Since the \emph{SOT-style} output consists of a single stream ($\nchn=1$), the execution time of \gls{MIMO-WER} is reasonable.


The \verb|asclite| tool is, according to its documentation\footnote{\url{https://github.com/usnistgov/SCTK/blob/master/doc/asclite.pod}}, able to handle all recognizer styles while performing a word-level matching ignoring speaker labels.
But in practice, it is computationally demanding and not all promoted features work as expected.
For example, the authors of \cite{Chen2020_ContinuousSpeechSeparation} had to pre-process the hypothesis transcriptions before they could apply the tool.



\section{Time-constrained WER}\label{sec:time-constrained-wer}

\glsreset{tcpWER}
All \gls{WER} metrics presented so far can match words as correct or substituted across arbitrary temporal distances, even though it is unlikely or impossible that the matched words stem from the same acoustic event (e.g., a speaker says common words like \enquote{the} and \enquote{and} many times during an hour-long session).
A human would recognize such a matching as implausible, so the metric should forbid such a matching as well.

To achieve this, we propose to incorporate a temporal constraint into the Levenshtein distance for \gls{WER} computation.
We specifically propose the \gls{tcpWER}, where the temporal constraint is introduced in the \gls{cpWER} by replacing the Levenshtein distance in \cref{eq:cp} with its time-constrained variant.

We assume that temporal annotations are present for each word in the form of a beginning and ending time.
We discuss later in \cref{sec:pseudo-word-level-annotations} how segment-level annotations can be used.
Words in the reference and hypothesis are only allowed to match (as correct or substituted) if they overlap or the gap between them is smaller than a collar $\collar$.
This is equivalent to setting the cost for a substitution and a correct match to $\costsub=\costcorr\geq\costins+\costdel$ where words are not allowed to match.

From the time constraint follows a straightforward optimization:
Cells in $\lm$ can be skipped where no words overlap since it is always valid to pick an insertion and deletion over a substitution or a correct match.
The skipped cells in \cref{eq:lev} form continuous regions in the upper right and bottom left of the Levenshtein matrix, so that only a diagonal band is left.

In the following, we describe how to deal with segment-level time annotations, then we discuss how to choose the size of the collar, followed by a more general discussion.


\begin{figure}
    \centering
    \input{tikz/pseudo-word-level}
    \caption{
        Visualization of the different pseudo-word-level annotation strategies. 
        The collar is visualized as gray boxes and kept short for better visualization.
        The character-based annotation strategy correlates best with the actual pronunciation time.
    }\label{fig:pseudo-word-level}
\end{figure}
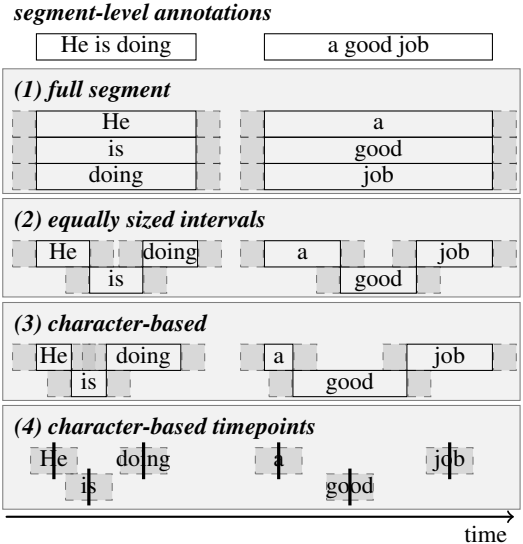

\subsection{Pseudo-word-level annotations}\label{sec:pseudo-word-level-annotations}

Word-level time-annotations are often not available, both for reference and hypothesis.
Obtaining them for datasets is expensive, error-prone and usually not deterministic.
Widely used single-speaker datasets like LibriSpech \cite{Panayotov2015_LibrispeechASRCorpus} and WSJ0/1 \cite{Garofalo2007_CsriWsj0Complete} do not provide them.
While a so-called forced alignment can be used to obtain annotations,
they are not reproducible and are not reliable for real recordings that contain overlapping speech and noise, like CHiME-6 \cite{Watanabe2020_CHiME6ChallengeTackling}.
On the hypothesis side, obtaining such detailed timing information has become difficult with the rise of attention-based models that provide no direct connection between input and output tokens. 

To overcome the need for precise word-level timing information, we propose to infer pseudo-word-level annotations from coarse segment-level annotations, see \cref{fig:pseudo-word-level}.
The simplest approach is (1) to use the full \emph{segment-level annotation} for each word.
This, however, does not represent the true word positions.
Using (1) for the hypothesis can be easily fooled: A system could predict a single segment that spans the whole recording length, e.g., by merging all recognized segments, which would make the \gls{tcpWER} equal to the \gls{cpWER} and render \gls{tcpWER} meaningless.
We conclude that annotations within a speaker must be non-overlapping for the hypothesis.

Instead, the true word boundaries can be roughly approximated by (2) \emph{equally sized intervals}, i.e., dividing each segment into number-of-words many intervals of equal size.
To incorporate differences in the time required to pronounce words, the (3) \emph{character-based} approximation chooses the segment length proportional to the number of characters in the word.
We assume that the number of characters correlates with the pronunciation length, which we confirmed for read English speech.
The average difference between the ground truth word length and the character-based approximation is, excluding examples with extreme annotation errors, below \SI{100}{ms} for both LibriSpeech and TIMIT.
Note that the approximation becomes less accurate for longer segments.
We recommend (3) for the reference.

The metric could be tricked when using (3) for the hypothesis by filling gaps between segments with single words.
This reintroduces implausible matchings that we aim to eliminate with the \gls{tcpWER}. 
We recommend to use (4) \emph{character-based timepoints}, i.e., the center points of (3), for the hypothesis.

While a phone-based approximation would be more accurate, it is in practice non-trivial to obtain and not unique.
We thus stick to the character-based approximation which can easily be computed for any transcript.



\subsection{Collar}
\begin{figure}
    \centering
    \input{tikz/word_distances}
    \vspace{-1.5em}
    \caption{
        Density plot of the gap sizes between pseudo-word-level annotations and ground-truth word-level annotations of each word for TIMIT and LibriSpeech.
        }
    \label{fig:distances}
\end{figure}
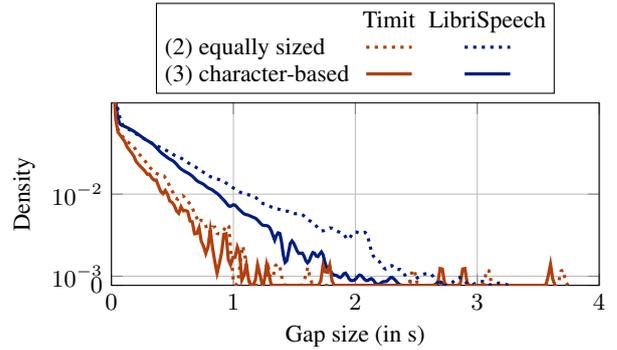

A collar $\collar>0$ is required to compensate inaccuracies in the (reference) segment-level annotations and approximation errors introduced by the pseudo-word-level annotations.
%
The reference annotations usually extend slightly over the actual pronunciation time.
Human annotators tend to over-estimate the length of the speech segments to make sure that all speech is covered.
For example, in TIMIT \cite{GarofoloJohnS1993_TIMITAcousticPhoneticContinuous} and LibriSpeech \cite{Panayotov2015_LibrispeechASRCorpus}, the recordings extend over the true speech activity by about \SI{400}{ms} and \SI{660}{ms} on average.\footnote{Actual speech activity determined with Kaldi \cite{Povey2011_KaldiSpeechRecognition}}
A similar pattern can be observed for simulated or re-recorded datasets since they are typically based on single-speaker datasets.
Additional issues can be introduced by the rerecording process, such as constant offsets induced by the playback hardware and sound propagation or a varying offset caused by a sample rate offset.
The collar should be chosen large enough to compensate errors introduced by these issues.

\subsubsection{Choosing a collar: pseudo-word-level approximation}\label{sec:eval-collar-lower-bound}

The collar should compensate the errors introduced by the pseudo-word-level approximation.
To determine which collar size compensates all these errors, we look at the gap sizes between the pseudo-word-level approximation and the true word boundaries\footnote{Determined with forced alignments using the Kaldi toolkit\cite{Povey2011_KaldiSpeechRecognition}}.
Their distribution is shown for the TIMIT \cite{GarofoloJohnS1993_TIMITAcousticPhoneticContinuous} and LibriSpeech \cite{Panayotov2015_LibrispeechASRCorpus} datasets in \cref{fig:distances} as a case study.
The distributions show large peaks at \SI{0}{s}, which means that for most words the pseudo-word-level annotations overlap with the true word boundaries with $\collar=0\,\si{s}$.
The character-based approximation, as expected, matches the true word boundaries better than the equally sized intervals.
Since a system that estimates words and timestamps correctly should have a \gls{WER} of zero, we obtain a lower bound of \SI{3.6}{s} and \SI{2.7}{s} for datasets that are based on TIMIT\footnote{The value for TIMIT is dominated by a single annotation error} and LibriSpeech, respectively.


\subsubsection{Choosing a collar: Approaching the desired WER}

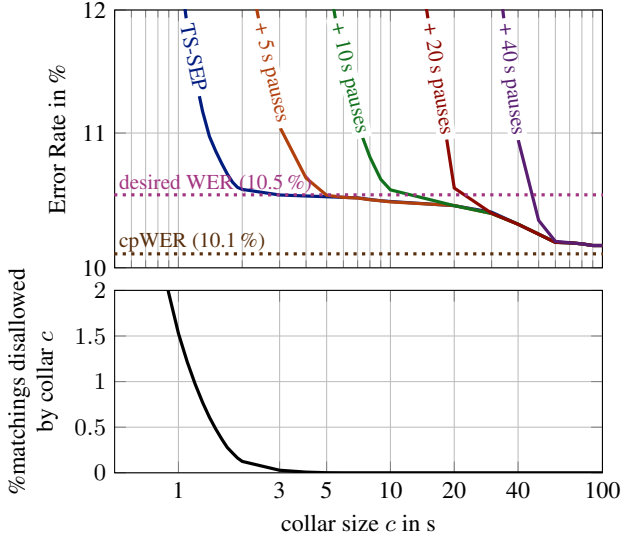
\begin{figure}
    \centering
    \input{tikz/tcpwer_over_collar}
    \caption{
        \emph{Top:} tcpWER over collar for the TS-SEP model on Libri-CSS. 
        The \enquote{desired WER} is determined with oracle word-level timestamps.
        \enquote{+ $n$\,\si{s} pauses} means that segments were artificially merged to include silence of $n$ seconds length.
        \emph{Bottom:} Proportion of matchings in the desired WER that would be disallowed by the collar. This value should be 0.
    }
    \label{fig:wer-vs-collar}
\end{figure}
For the following investigation, we use the recognition result for the Libri-CSS \cite{Chen2020_ContinuousSpeechSeparation} dataset obtained with a target-speaker separation (TS-SEP) \cite{Boeddeker2023_TSSEPJointDiarization} model
followed by the base model from Whisper \cite{Radford2022_RobustSpeechRecognition} as a single-speaker speech recognizer which delivers word boundaries needed for our analysis.
This system generates diarization-style output.

The top of \cref{fig:wer-vs-collar} shows the \gls{tcpWER} over the collar.
The \gls{tcpWER} naturally approaches the \gls{cpWER} for an increasing collar $\collar$ since the \gls{tcpWER} = \gls{cpWER} for $\collar\rightarrow\infty$.
The \gls{tcpWER} decreases quickly for small $\collar$ and is roughly constant in the range of \SIrange{3}{10}{s}, which agrees with \cref{sec:eval-collar-lower-bound}.

The collar should be chosen such that it forbids implausible matchings but does not forbid matching of words that actually stem from the same acoustic event.
We first define the \enquote{desired WER} and the desired matching, where words are only identified as correct when they temporally overlap (without a collar) using word-level timestamps.
Reference word-level timestamps are obtained using forced alignments (including an offset compensation for the playback delay) and for the hypothesis using Whipser \gls{ASR}.
The \gls{cpWER} and the \enquote{desired WER} are shown in \cref{fig:wer-vs-collar} as dashed lines.
It should be noted that the difference between the \gls{cpWER} and the desired WER is relatively small since the overall performance of the model is relatively good.
The \gls{tcpWER} is close to the desired \gls{WER} for a collar between \SI{3}{s} and \SI{5}{s} which means that this collar size successfully compensate errors introduced by the reference annotation but still forbids temporally implausible matchings.

The bottom of \cref{fig:wer-vs-collar} shows the amount of matchings in the \enquote{desired WER} (after offset correction) that would be disallowed by the collar (without offset correction and with pseudo-word-level annotations).
This curve includes the errors from \cref{fig:distances}, from the estimation of the pseudo-word-level annotations and offset errors introduced by the simulation and recording process of Libri-CSS.
Its value should be 0 for the chosen collar value, which gives us a lower bound of \SI{4}{\second}.

To investigate the impact of silence or long pauses in a system output on the metric, we artificially merge segments to include such pauses, denoted by \enquote{+ $n$\,\si{s} pauses} in \cref{fig:wer-vs-collar} for pauses up to $n$ seconds.
Large silences should negatively impact the metric which is not reflected in the \gls{cpWER} but the \gls{tcpWER} increases more silence.
Segments that include large silent regions are penalized because the quality of the pseudo-word-level annotation decreases with increasing amounts of silence.

We argue that short silences in the range of a few seconds do not hurt downstream systems and should be allowed in the system output and the reference.
An allowed silence length of \SI{5}{s} (visualized as \enquote{+\SI{5}{s} pauses} in \cref{fig:wer-vs-collar}) seems reasonable here since it is unlikely that a real system includes much longer silences and the errors in the reference are typically smaller.
A collar of \SI{5}{\second} is thus a reasonable choice for the majority of systems.
When applying the \gls{tcpWER} to other datasets, the user should be aware of the issues mentiond above and adjust the collar value accordingly.

\begin{table}[t]
    \setlength{\tabcolsep}{5pt}
    \centering
    \caption{cpWER and tcpWER on different datasets}
    \begin{tabular}{l@{~}c@{~~~~~~}lSSHHHH}
        \toprule
        Dataset & \clap{(\#spks)}    & Model     & {cpWER (\%)} & {tcpWER (\%)} & {Relative Difference} & MIMO-WER & asclite & DER \\
        \midrule
        Libri-CSS & (8)   & TS-SEP    & 10.1  & 10.5 & 3.8 \\
        CHiME-6 & (4)     & baseline  & 62.4 & 66.6 & 8.3 \\
        DiPCo & (4)      & baseline  & 56.8 & 61.7 & 10.1 \\
        Mixer 6 & (2)     & baseline  & 20.3 & 20.4 & 1.6 \\
        \bottomrule
    \end{tabular}
    \label{tab:scoring}
\end{table}

\subsection{Relation to other metrics}

The idea to use a temporal constraint has appeared before for single utterance recognition, e.g., in \verb|sclite| and \cite{Morris2004_WERRILMER}, but required exact word-level annotations.
Due to the issues discussed in \cref{sec:pseudo-word-level-annotations}, it is often impossible to obtain such annotations for long-form recognition, and thus has been ignored in the past for this scenario.
The time-based cost model in \verb|sclite| incorporates the distance between words into the cost for a substitution, which makes the model prefer a deletion and insertion over a substitution when words do not overlap.
It has, however, no notion of a collar and the constraint is less strict.

For evaluation of long-form speech recognition, the \verb|asclite| \cite{Fiscus2006_MultipleDimensionLevenshtein} tool uses a temporal constraint, but mainly for speedup reasons.
It has no clear notion of a collar and is not designed to improve the quality of the matching.






\subsection{Scoring results}\label{sec:scoring-results}

Scoring results for the CHiME-7 baseline system on the CHiME-6 \cite{Watanabe2020_CHiME6ChallengeTackling}, DiPCo \cite{Segbroeck2019DiPCo} and Mixer 6 \cite{Brandschain2010_Mixer6} datasets, and with the TS-SEP model on the Libri-CSS dataset are shown in \cref{tab:scoring}.
We focus on Diarizaiton-style recognizers here, as they are required for the CHiME-7 challenge.
We thus only provide results for \gls{cpWER} and \gls{tcpWER} and refer to \cite{vonNeumann2023_WordErrorRate} for \gls{ORC-WER} and MIMO-WER.
The \gls{tcpWER} is computed with a collar of $\collar=5\,\si{s}$.
As expected, the \gls{tcpWER} is always larger than the \gls{cpWER}.
Especially for poor recognition performance, as on CHiME-6 and DiPCo, we can observe larger differences.

We observed that higher \glspl{WER} (more wrongly transcribed words) lead to a higher number of unreasonable matchings.
The number of unreasonable matchings is further increased by longer inactivity of a single speaker between utterances.
The number of speakers in a recording is a proxy for the amount of silence which is reflected by larger differences between \gls{cpWER} and \gls{tcpWER} in \cref{tab:scoring}.


\subsection{Profiling}\label{sec:profiling}

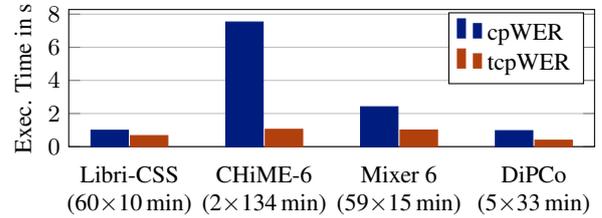
\begin{figure}
    \centering
    \input{tikz/profiling}
    \caption{
        Execution time  of cpWER and tcpWER on different datasets.
        The number of recordings in the dataset and the average recording length is given in parenthesis.
    }
    \label{fig:profiling}
\end{figure}

The profiling results of \gls{cpWER} compared with \gls{tcpWER} are shown in \cref{fig:profiling} for LibriCSS, CHiME-6, DiPCo and Mixer 6.
Execution times were measured on a single core of a an Intel Core i7-13700K processor, averaged over 10 runs.
The execution times of both \glspl{WER} are in an acceptable range, but the pruning employed in \gls{tcpWER} leads to a decrease in execution time with increasing total lengths of the evaluated recordings.

\section{Conclusions}

This paper presents the MeetEval toolkit which implements a variety of Word Error Rate metrics for the evaluation of meeting transcription systems.
We propose to incorporate a temporal constraint into the WER computation to improve the matching between words in the reference and hypothesis and prevent unrealistic matchings.
We show that the temporal constraint with a reasonably chosen collar leads to realistic WERs.
A comprehensive intuition is given for how to choose a reasonable collar.

\section{Outlook}

While the time constraint was presented only in the tcpWER as an extension to the cpWER in this work, it can be naively incorporated into the ORC-WER and MIMO-WER without major conceptual modifications, resulting in the tcORC-WER and tcMIMO-WER.
The naive implementation, however, does not improve the runtime complexity.
The time constraint allows for a significant reduction in runtime complexity by pruning the search space.
This is left for future work.

\section{Acknowledgements}

Christoph Boeddeker was funded by Deutsche Forschungsgemeinschaft (DFG), project no.\ 448568305. 
Computational resources were provided by BMBF/NHR/PC2.
We thank Samuele Cornell for kindly providing the transcription results of the CHiME-7 baseline system.

\bibliographystyle{IEEEtran}
\balance
\bibliography{references}

\end{document}

%% file: macros.tex
\usepackage{amsmath}
\usepackage{amssymb}
\usepackage{bbm}
\usepackage{physics}
\usepackage{siunitx}
\usepackage{trfsigns}
\usepackage{pifont}
\usepackage{etoolbox}
\usepackage{multirow}
\usepackage{csquotes}
\usepackage[skins]{tcolorbox}
\usepackage{cite}   
\usepackage{hyperref}
\usepackage[capitalize]{cleveref}
\usepackage{balance} 
\usepackage{booktabs}
\usepackage{listings}
\usepackage{xifthen}    
\usepackage{subcaption}

\usepackage{url}

\let\oldcite\cite
\renewcommand{\cite}[1]{%
\ifthenelse{\isempty{#1}}%
{\inred{[cite!]}}%
{\oldcite{#1}}%
}

\newcolumntype{H}{>{\setbox0=\hbox\bgroup}c<{\egroup}@{}}   

\newcommand*{\thl}{\fontseries{b}\selectfont}
\robustify\thl
\robustify\fontseries

\newcommand{\inred}[1]{\textcolor{red}{#1}}

\makeatletter
\renewcommand\subsection{\@startsection{subsection}{2}{\z@}%
                                     {-1.5ex\@plus -1ex \@minus -.2ex}%
                                     {1.15ex \@plus .2ex}%
                                     {\normalfont\large\bfseries}}
\renewcommand\subsubsection{\@startsection{subsubsection}{3}{\z@}%
                                     {-2.25ex\@plus -1ex \@minus -.2ex}%
                                     {0.5ex \@plus .2ex}%
                                     {\normalfont\normalsize\bfseries}}
\makeatother

%% file: tikz_style.tex
\usepackage{tikz}
\usepackage{pgfplots}
\pgfplotsset{compat=1.16}
\usepgfplotslibrary{groupplots}
\usetikzlibrary{positioning,arrows,matrix,fit,calc,patterns,chains,scopes,shapes.multipart,decorations,decorations.markings,backgrounds}
\usetikzlibrary{intersections}

\definecolor{palette-1}{HTML}{001C7F}    
\definecolor{palette-2}{HTML}{B1400D}    
\definecolor{palette-3}{HTML}{12711C}    
\definecolor{palette-4}{HTML}{8C0800}    
\definecolor{palette-5}{HTML}{591E71}    
\definecolor{palette-6}{HTML}{592F0D}    
\definecolor{palette-7}{HTML}{A23582}    
\definecolor{palette-8}{HTML}{3C3C3C}    
\definecolor{palette-9}{HTML}{B8850A}    
\definecolor{palette-10}{HTML}{006374}      

\tikzset{
    line/.style={draw,black,thick,rounded corners=1mm,line cap=round},
    noshortarrow/.style={line,->},
    arrow/.style={noshortarrow,shorten >=.3mm},
    doublearrow/.style={arrow,<->, shorten <=.3mm},
    box/.style={draw,black,thick,minimum height=3em,text depth=0.25ex,rounded corners=3,fill=white},
    nopadding/.style={minimum height=0,inner sep=1mm},
    signalbox/.style={draw,black,thin,rounded corners=1mm,minimum width=7mm, minimum height=4mm,inner sep=0},
    pbox/.style={box,fill=black!10},
    backgroundbox/.style={inner xsep=3mm, inner ysep=1mm, draw, dashed, rounded corners,fill=orange!10},
    branch/.style={inner sep=0.3mm,circle,fill=black},
    operator/.style={draw,circle,black,rounded corners,inner sep=0,fill=white},
    vertex/.style={draw,ultra thin,circle,black,rounded corners,inner sep=0.6mm,fill=gray,fill opacity=0.5},
    edge/.style={line,very thick,line cap=butt},
    pattern1/.style={pattern=north west lines,pattern color=palette-1},
    pattern2/.style={pattern=north east lines,pattern color=palette-2},
    pattern3/.style={pattern=crosshatch,pattern color=palette-3},
    buswidth/.style={path picture={\draw[black,-] (path picture bounding box.south west) -- (path picture bounding box.north east);}}
}

\tikzset{
    word/.style={
        execute at begin node={\vphantom{Ag}},
        execute at end node={\vphantom{Ag}},
        inner sep=0.05em,
        font = {\tiny},
        line width=0,
    },
    spkOne/.style={fill=palette-1, text=white},
    spkTwo/.style={fill=palette-2, text=white},
    spkThree/.style={fill=palette-3, text=white},
    outputScope/.style={x=1em, y=-.7em},
    outputBoundingBox/.style={draw=black, fill=gray!40},
    tableSep/.style={line,draw=gray!50},
}

%% file: symbols.tex
\newcommand\defm[2]{\expandafter\newcommand{#1}{\ensuremath{#2}}} 

\defm\Loss{\mathcal{L}}    

\defm\tref{\mathcal{R}}
\defm\thyp{\mathcal{H}}

\defm\nspk{I}
\defm\nchn{J}
\defm\nutt{U}

\defm\iref{r}
\defm\ihyp{h}
\defm\ispk{i}
\defm\ichn{j}
\defm\iutt{u}

\defm{\lev}{\mathrm{lev}}
\defm\bigO{\mathcal{O}}

\defm\collar{c}

\defm\aref{i}
\defm\ahyp{j}

\defm\costcorr{C_\text{corr}}
\defm\costsub{C_\text{sub}}
\defm\costdel{C_\text{del}}
\defm\costins{C_\text{ins}}
\defm\costcorrsub{C_\text{corr/sub}}
\defm\lm{L}
\defm\li{r}
\defm\lj{h}

\defm\reals{\mathbb{R}}

%% file: glossaries.tex
\usepackage[acronym,shortcuts]{glossaries}
\glsdisablehyper    

\newacronym{WER}{WER}{Word Error Rate}
\newacronym{cpWER}{cpWER}{Concatenated minimum-Permutation WER}
\newacronym{ORC-WER}{ORC-WER}{Optimal Reference Combination WER}
\newacronym{MIMO-WER}{MIMO-WER}{\inred{XXXXXXXX} WER}
\newacronym{DER}{DER}{Diarization Error Rate}
\newacronym{tcpWER}{tcpWER}{Time-Constrained minimum-Permutation WER}
\newacronym{ASR}{ASR}{Automatic Speech Recogniton}

%% file: tikz/scenario.tex

\begin{tikzpicture}[
    hlabel/.style={font={\footnotesize},inner sep=0,anchor=south west,yshift=.4em}
]
    \begin{scope}[xshift=-5em, yshift=-5em, local bounding box=outputScope, outputScope]
        \node[word,anchor=west] at (-1.5em,0) (hyp1) {$\thyp_1$:};
            \node[word,anchor=west] at (-1.5em,1) (hyp2) {$\thyp_2$:};
            \node[word,anchor=west] at (-1.5em,2) (hyp3) {$\thyp_3$:};
            \node[word,spkOne,anchor=west] (the) at (0, 0) {The};
            \node[word,spkOne,anchor=west] (quick) at ($(the.east)+(0.2,0)$) {quick};
            \node[word,spkOne,anchor=west] (brown) at ($(quick.east)+(0.6,0)$) {brown};
            \node[word,spkOne,anchor=west] (fox) at ($(brown.east)+(0.2,0)$) {fox};
            \node[word,spkTwo,anchor=west] (over) at ($(quick.east)+(0.1,1)$) {over};
            \node[word,spkThree, anchor=east] (lazy) at ($(brown.west)+(-0.1,2)$) {lazy};
            \node[word,spkTwo, anchor=east] (jumps) at ($(lazy.west)+(-0.2,-1)$) {jumps};
            \node[word,spkThree] (dog) at ($(brown.east)!1/2!(fox.west)+(0,2)$) {dog};
    \end{scope}
    \node[hlabel] (labelDia) at (outputScope.north west) {\emph{Grouped by speaker}};
    \begin{pgfonlayer}{background}
    \node[outputBoundingBox, fit=(the) (fox) (dog)(hyp1)] (dia) {};
    \end{pgfonlayer}
    
    \begin{scope}[yshift=-9.5em, local bounding box=outputScope, outputScope]
            \coordinate (scopeOrigin) at (0, 0);
            \node[word,spkOne] (the) at ($(scopeOrigin-|the)$) {The};
            \node[word,anchor=east] at ($(the.west)+(-0.2em,0)$) (hyp1) {$\thyp_1$:};
            \node[word,spkOne] (quick) at ($(scopeOrigin-|quick)+(0,0)$) {quick};
            \node[word,spkOne] (brown) at ($(scopeOrigin-|brown)+(0,1)$) {brown};
            \node[word,spkOne] (fox) at ($(scopeOrigin-|fox)+(0,1)$) {fox};
            \node[word,spkTwo] (jumps) at ($(scopeOrigin-|jumps)+(0,1)$) {jumps};
            \node[word,anchor=east] at ($(hyp1.east|-jumps)$) (hyp2) {$\thyp_2$:};
            \node[word,spkTwo] (over) at ($(scopeOrigin-|over)+(0,0)$) {over};
            \node[word,spkThree] (lazy) at ($(scopeOrigin-|lazy)+(0,1)$) {lazy};
            \node[word,spkThree] (dog) at ($(scopeOrigin-|dog)+(0,0)$) {dog};
    \end{scope}
    \node[hlabel] (labelCSS) at (outputScope.north west) {\emph{$\leq\nchn$ overlap-free channels}};
    \begin{pgfonlayer}{background}
        \node[outputBoundingBox, fit=(the) (fox) (dog) (hyp2)] (css) {};
    \end{pgfonlayer}
    
    \begin{scope}[yshift=-13em, local bounding box=outputScope, outputScope, anchor=west]
        \coordinate (scopeOrigin) at (0, 0);
        \node[word,spkOne,anchor=center] (the) at ($(scopeOrigin-|the)$) {The};
        \node[word,anchor=east] at ($(the.west)+(-0.2em,0)$) (hyp1) {$\thyp_1$:};
        \node[word,spkOne] (quick) at ($(the.east)+(0.1,0)$) {quick};
        \node[word,spkOne] (brown) at ($(quick.east)+(0.1,0)$) {brown};
        \node[word,spkOne] (fox) at ($(brown.east)+(0.1,0)$) {fox};
        \node[word,spkTwo] (jumps) at ($(fox.east)+(0.1,0)$) {jumps};
        \node[word,spkTwo] (over) at ($(jumps.east)+(0.1,0)$) {over};
        \node[word,spkThree] (lazy) at ($(over.east)+(0.1,0)$) {lazy};
        \node[word,spkThree] (dog) at ($(lazy.east)+(0.1,0)$) {dog};
    \end{scope}
    \node[hlabel] (labelSOT) at (outputScope.north west) {\emph{Serialized, speaker ordered}};
    \begin{pgfonlayer}{background}
    \node[outputBoundingBox, fit=(the) (fox) (dog) (hyp1)] (sotSpk) {};
    \end{pgfonlayer}
    
    \begin{scope}[yshift=-15.5em, local bounding box=outputScope, outputScope, anchor=west]
        \coordinate (scopeOrigin) at (0, 0);
        \node[word,spkOne,anchor=center] (the) at ($(scopeOrigin-|the)$) {The};
        \node[word,anchor=east] at ($(the.west)+(-0.2em,0)$) (hyp1) {$\thyp_1$:};
        \node[word,spkTwo] (jumps) at ($(the.east)+(0.1,0)$) {jumps};
        \node[word,spkOne] (quick) at ($(jumps.east)+(0.1,0)$) {quick};
        \node[word,spkThree] (lazy) at ($(quick.east)+(0.1,0)$) {lazy};
        \node[word,spkTwo] (over) at ($(lazy.east)+(0.1,0)$) {over};
        \node[word,spkOne] (brown) at ($(over.east)+(0.1,0)$) {brown};
        \node[word,spkThree] (dog) at ($(brown.east)+(0.1,0)$) {dog};
        \node[word,spkOne] (fox) at ($(dog.east)+(0.1,0)$) {fox};
    \end{scope}
    \node[hlabel] at (outputScope.north west) {\emph{Serialized, temporal ordered}};
    \begin{pgfonlayer}{background}
    \node[outputBoundingBox, fit=(the) (fox) (dog) (hyp1)] (sotSeg) {};
    \end{pgfonlayer}

    \node[align=center,anchor=east] (asrDia) at ($(dia.west) + (-2em,0)$) {Diarization-style\\Recognizer};

    \node[align=center,anchor=center] (asrCSS) at ($(css-|asrDia)$) {CSS-style\\Recognizer};

    \node[align=center,anchor=center] (asrSOT) at ($(sotSpk.west-|asrDia)!1/2!(sotSeg.west-|asrDia) + (0em,0)$) {SOT-style\\Recognizer};

    \node[align=center,anchor=west] (cpwer) at ($(dia-|sotSeg.east) + (3em,0)$) {cpWER\\tcpWER};
    \node[align=center,anchor=center,yshift=1mm] (csswer) at ($(css-|cpwer)$) {ORC-WER\\MIMO-WER\\asclite};
    \node[align=center,anchor=center] (sotwer) at ($(sotSpk-|cpwer)!1/2!(sotSeg-|cpwer)$) {MIMO-WER\\(ORC-WER)};

    \node (out) at ($(sotSeg|-dia) + (0,5em)$) {Hypothesis transcript $\thyp$};
    \node[box, fill=palette-1!50] (asr) at (out-|asrDia) {Speech Recognizer};

    \node[left=4em of asr,fill stretch image=images/signal1,minimum width=3em, minimum height=2em] (input) {};
    \node[fill stretch image=images/signal2,minimum width=2em,xshift=1em, minimum height=2em] (wav2) at (input) {};
    \node[fill stretch image=images/signal3,minimum width=2.5em,xshift=3em, minimum height=2em] (wav3) at (input) {};

    \node[align=center,yshift=1mm] at (input|-css) {At most $\nchn$ \\ simultaneously over- \\ lapping speakers};

    \node[box,fill=palette-3!50] (metric) at (out-|cpwer) {Metric};
    \node[right=2em of metric] (reference) {Reference Transcript $\tref$};

    \node[anchor=north west,inner sep=0] (label) at (sotSeg.west|-asr.south) {\bfseries\emph{example}};
    \node[anchor=north west,inner sep=0] (label) at (csswer.west|-asr.south) {\bfseries\emph{recommended}};

    \draw[arrow] (wav3) -- (asr);
    \draw[arrow] (asr) -- (out);
    \draw[arrow] (out) -- (metric);
    \draw[arrow] (reference) -- (metric);

    \coordinate (sep0) at ($(label.south)!0.5!(labelDia.north)$);
    \draw[tableSep] (sep0-|input.west) -- (sep0-|reference.east);
    \coordinate (sep1) at ($(asrDia.south)!0.5!(labelCSS.north)$);
    \draw[tableSep] (sep1-|input.west) -- (sep1-|reference.east);
    \coordinate (sep2) at ($(asrCSS.south)!0.5!(labelSOT.north)$);
    \draw[tableSep] (sep2-|input.west) -- (sep2-|reference.east);

    \begin{pgfonlayer}{background}
        \node[fit={(asrSOT)(asrDia)(asr)(asr|-sotSeg.south)},draw,fill=palette-1!10]{};
        \node[fit={(sotSeg)(dia)(out)(asr.north-|sotSeg)},draw]{};
        \node[fit={(metric)(cpwer)(sotwer)(metric|-sotSeg.south)},draw,fill=palette-3!10]{};
    \end{pgfonlayer}
    
\end{tikzpicture}

%% file: tikz/pseudo-word-level.tex
\begin{tikzpicture}[xscale=.5]
    \def\gap{8.5mm}
    \tikzset{
        segment/.style={
            draw=black,minimum height=1.1em,anchor=west,
            execute at begin node={\vphantom{Ag}},
            execute at end node={\vphantom{Ag}},
            font = {\small},
            inner sep=0,
        },
        point/.style={
            anchor=west,
            inner sep=0,
            minimum height=1.1em,
        },
        pointtext/.style={
            execute at begin node={\vphantom{Ag}},
            execute at end node={\vphantom{Ag}},
            font = {\small},
            inner sep=0,
            anchor=center,
            minimum height=1.1em,
        },
        collar/.style={segment,dashed,fill=gray!50,opacity=0.5,minimum width=3mm}
    }
    \begin{scope}[local bounding box=original]
        \node[segment,minimum width=2.1cm] (segment1) {He is doing};
        \node[segment,minimum width=3cm,xshift=3cm] (segment2) {a good job};
    \end{scope}
    \node[anchor=south west,xshift=-4mm] at (original.north west) (label) {\textbf{\emph{segment-level annotations}}};
    \begin{scope}[shift={($(original.south-|0,0)+(0,-\gap)$)},local bounding box=fullsegment]
        \node[segment,minimum width=2.1cm] (s11) {He};
        \node[segment,minimum width=2.1cm,yshift=-1.1em] (s12) {is};
        \node[segment,minimum width=2.1cm,yshift=-2.2em] (s13) {doing};
        \node[segment,minimum width=3cm,xshift=3cm] (s21) {a};
        \node[segment,minimum width=3cm,xshift=3cm,yshift=-1.1em] (s22) {good};
        \node[segment,minimum width=3cm,xshift=3cm,yshift=-2.2em] (s23) {job};

        \foreach \i in {s11,s12,s13,s21,s22,s23}{
            \node[collar,anchor=west] at (\i.east) {};
            \node[collar,anchor=east] at (\i.west) {};
        }
        \node[anchor=south west] at (s11.north-|label.west) (label) {\textbf{\emph{(1) full segment}}};
    \end{scope}
    \begin{pgfonlayer}{background}
        \node[fit={(fullsegment)}, inner sep=0.5mm,fill=gray!10,draw=gray] {};
    \end{pgfonlayer}

    \begin{scope}[shift={($(fullsegment.south-|0,0)+(0,-\gap)$)},local bounding box=equispaced]
        \node[segment,minimum width=.7cm] (s11) {He};
        \node[segment,minimum width=.7cm,xshift=.7cm,yshift=-1.1em] (s12) {is};
        \node[segment,minimum width=.7cm,xshift=1.4cm] (s13) {doing};
        \node[segment,minimum width=1cm,xshift=3cm] (s21) {a};
        \node[segment,minimum width=1cm,xshift=4cm,yshift=-1.1em] (s22) {good};
        \node[segment,minimum width=1cm,xshift=5cm] (s23) {job};

        \foreach \i in {s11,s12,s13,s21,s22,s23}{
            \node[collar,anchor=west] at (\i.east) {};
            \node[collar,anchor=east] at (\i.west) {};
        }
        \node[anchor=south west] at (s11.north-|label.west) (label) {\textbf{\emph{(2) equally sized intervals}}};
    \end{scope}
    \begin{pgfonlayer}{background}
        \node[fit={(equispaced)}, inner sep=0.5mm,fill=gray!10,draw=gray] {};
    \end{pgfonlayer}

    \begin{scope}[shift={($(equispaced.south-|0,0)+(0,-\gap)$)},local bounding box=char]
        \node[segment,minimum width=.46cm] (s11) {He};
        \node[segment,minimum width=.46cm,xshift=.46cm,yshift=-1.1em] (s12) {is};
        \node[segment,minimum width=.98cm,xshift=0.92cm] (s13) {doing};
        \node[segment,minimum width=0.375cm,xshift=3cm] (s21) {a};
        \node[segment,minimum width=1.5cm,xshift=3.375cm,yshift=-1.1em] (s22) {good};
        \node[segment,minimum width=1.125cm,xshift=4.875cm] (s23) {job};

        \foreach \i in {s11,s12,s13,s21,s22,s23}{
            \node[collar,anchor=west] at (\i.east) {};
            \node[collar,anchor=east] at (\i.west) {};
        }

        \node[anchor=south west] at (s11.north-|label.west) (label) {\textbf{\emph{(3) character-based}}};
    \end{scope}
    \begin{pgfonlayer}{background}
        \node[fit={(char)}, inner sep=0.5mm,fill=gray!10,draw=gray] {};
    \end{pgfonlayer}

    \begin{scope}[shift={($(char.south-|0,0)+(0,-\gap)$)},local bounding box=charpoint]
        \node[point,minimum width=.460cm] (s11) {};
        \node[point,minimum width=.46cm,xshift=.46cm,yshift=-1.1em] (s12) {};
        \node[point,minimum width=.98cm,xshift=0.92cm] (s13) {};
        \node[point,minimum width=0.375cm,xshift=3cm] (s21) {};
        \node[point,minimum width=1.5cm,xshift=3.375cm,yshift=-1.1em] (s22) {};
        \node[point,minimum width=1.125cm,xshift=4.875cm] (s23) {};

        \foreach \i in {s11,s12,s13,s21,s22,s23}{
            \node[collar,anchor=west] at (\i.center) {};
            \node[collar,anchor=east] at (\i.center) {};
        }

        \foreach \i in {s11,s12,s13,s21,s22,s23}{
            \draw[black,very thick] ($(\i.south)+(0,-0.2em)$) -- ($(\i.north)+(0,0.2em)$);
        }

        \node[pointtext] at (s11.center) {He};
        \node[pointtext] at (s12.center) {is};
        \node[pointtext] at (s13.center) {doing};
        \node[pointtext] at (s21.center) {a};
        \node[pointtext] at (s22.center) {good};
        \node[pointtext] at (s23.center) {job};
        \node[anchor=south west] at (s11.north-|label.west) (label) {\textbf{\emph{(4) character-based timepoints}}};

        \node[inner sep=0] at ($(fullsegment.east|-s11)+(-1pt,0)$) {};
    \end{scope}
    \begin{pgfonlayer}{background}
        \node[fit={(charpoint)}, inner sep=0.5mm,fill=gray!10,draw=gray]  {};
    \end{pgfonlayer}

    \draw[arrow] ([yshift=-1.5mm]charpoint.south west) -- ([yshift=-1.5mm]charpoint.south east) node[below,anchor=north east]{time};
\end{tikzpicture}

%% file: tikz/word_distances.tex
\begin{tikzpicture}
    \begin{axis}[
        width=8cm,
        height=4cm,
        xmajorgrids,
        xmin=0, xmax=4,
        xminorgrids,
        ylabel=Density,
        xlabel=Gap size (in \si{\second}),
        ymajorgrids,
        ymin=0, ymax=0.02,
        yminorgrids,
        scaled y ticks=false,
        ytick={0,0.001,0.01},
        yticklabels={
          \(\displaystyle {0}\),
          \(\displaystyle {10^{-3}}\),
          \(\displaystyle {10^{-2}}\),
        }
    ]
    \addplot [very thick, palette-1]
table {%
-0.064657514964756 0.012465589417546
-0.0504802267846123 0.0162886772689661
-0.0363029386044687 0.0191760408034872
-0.022125650424325 0.0211310274313425
-0.00794836224418135 0.0221613791480135
0.0062289259359623 0.0222873450690638
0.020406214116106 0.0215690959784644
0.0345835022962496 0.0202004366950943
0.0487607904763933 0.0187230152808882
0.0629380786565369 0.0178675153710495
0.0771153668366806 0.0176036334721877
0.0912926550168243 0.0175120712511631
0.105469943196968 0.0174360144613839
0.119647231377112 0.0173546892701558
0.133824519557255 0.0172644544809163
0.148001807737399 0.0171640248322767
0.162179095917543 0.0170592430299818
0.176356384097686 0.0169497274559773
0.19053367227783 0.0168258834986577
0.204710960457974 0.0166907352627797
0.218888248638117 0.0165579931866225
0.233065536818261 0.0164262591360797
0.247242824998404 0.0162868718607202
0.261420113178548 0.0161462120927512
0.275597401358692 0.0160179370265175
0.289774689538836 0.0159033175556753
0.303951977718979 0.0157933991483551
0.318129265899123 0.0156830812726955
0.332306554079266 0.015561663724598
0.34648384225941 0.0154075702740853
0.360661130439554 0.0152172930808684
0.374838418619697 0.0150111882319604
0.389015706799841 0.0148019472441951
0.403192994979985 0.0146007782493831
0.417370283160128 0.0144326504584531
0.431547571340272 0.0142956038967476
0.445724859520416 0.0141552532980716
0.459902147700559 0.0140040822152114
0.474079435880703 0.0138727210279535
0.488256724060847 0.0137711552867112
0.50243401224099 0.0136650321087868
0.516611300421134 0.0135203494654743
0.530788588601278 0.0133262832238353
0.544965876781421 0.0130985621444528
0.559143164961565 0.0128876037361818
0.573320453141709 0.0127300815856526
0.587497741321852 0.0126071296552224
0.601675029501996 0.0124939131866019
0.61585231768214 0.0123897147324875
0.630029605862283 0.0122875851282619
0.644206894042427 0.0121679508277294
0.658384182222571 0.0120218613436952
0.672561470402714 0.0118494874523878
0.686738758582858 0.0116576299838732
0.700916046763002 0.0114783746510684
0.715093334943145 0.0113477542838036
0.729270623123289 0.0112595483483795
0.743447911303432 0.0111744489418572
0.757625199483576 0.0110582234159794
0.77180248766372 0.0109020366819543
0.785979775843863 0.0107397627055136
0.800157064024007 0.010627838589503
0.814334352204151 0.0105552374853013
0.828511640384294 0.010449060360888
0.842688928564438 0.0103017999727326
0.856866216744582 0.0101661671693614
0.871043504924725 0.0100331325532434
0.885220793104869 0.00985570362972199
0.899398081285013 0.00964771063712208
0.913575369465156 0.00945062286884706
0.9277526576453 0.00925604637142127
0.941929945825444 0.00902827152556674
0.956107234005587 0.00878391157741914
0.970284522185731 0.00864823852977997
0.984461810365875 0.00871549948052162
0.998639098546018 0.00883362177742067
1.01281638672616 0.00880458001329433
1.02699367490631 0.00861725766236108
1.04117096308645 0.00839862360292772
1.05534825126659 0.00822391279483576
1.06952553944674 0.00804878172877187
1.08370282762688 0.00787576309665689
1.09788011580702 0.00777635730391945
1.11205740398717 0.007735435220075
1.12623469216731 0.00767793420527
1.14041198034746 0.00758780485769467
1.1545892685276 0.00750329718659563
1.16876655670774 0.00743462488287993
1.18294384488789 0.00733710176985112
1.19712113306803 0.00717930971247337
1.21129842124817 0.00697070263533211
1.22547570942832 0.00675144426308738
1.23965299760846 0.00660050278434546
1.2538302857886 0.00655704929581271
1.26800757396875 0.00655443066303903
1.28218486214889 0.00651309198357777
1.29636215032904 0.00635580144579717
1.31053943850918 0.00599630762622288
1.32471672668932 0.00546187731710494
1.33889401486947 0.00495912013315418
1.35307130304961 0.00547853666069931
1.36724859122975 0.0057609395818467
1.3814258794099 0.00567939920620439
1.39560316759004 0.00470422350171683
1.40978045577018 0.00329992800533556
1.42395774395033 0.00251084785013967
1.43813503213047 0.00270511166857228
1.45231232031062 0.00365204994337225
1.46648960849076 0.00459142112928295
1.4806668966709 0.00487110669990534
1.49484418485105 0.00461630979362422
1.50902147303119 0.0042011912688615
1.52319876121133 0.00375928718455733
1.53737604939148 0.00349572208308945
1.55155333757162 0.00352721525680318
1.56573062575176 0.00353630508067262
1.57990791393191 0.00319802152019106
1.59408520211205 0.00269493977300378
1.6082624902922 0.00246411592347168
1.62243977847234 0.00268963811760293
1.63661706665248 0.00320297727996102
1.65079435483263 0.0035932755202622
1.66497164301277 0.00352863012070554
1.67914893119291 0.00321464693059854
1.69332621937306 0.00315201672168578
1.7075035075532 0.00344227088811713
1.72168079573335 0.00366554893818757
1.73585808391349 0.00333882288647357
1.75003537209363 0.00255172251292167
1.76421266027378 0.00195248606132092
1.77838994845392 0.00180014263840199
1.79256723663406 0.00166923867827598
1.80674452481421 0.00134857860960676
1.82092181299435 0.00112993757323215
1.83509910117449 0.00111233586216932
1.84927638935464 0.00106512750367001
1.86345367753478 0.000917758109971886
1.87763096571493 0.000856547471633488
1.89180825389507 0.0010185373189874
1.90598554207521 0.0012778792247108
1.92016283025536 0.00134088565476342
1.9343401184355 0.00108137661163055
1.94851740661564 0.000718264719179422
1.96269469479579 0.000526885283357204
1.97687198297593 0.000515982118773631
1.99104927115607 0.000574833293784951
2.00522655933622 0.000719991111028954
2.01940384751636 0.000960465392402997
2.03358113569651 0.00113062126948986
2.04775842387665 0.00107770311145192
2.06193571205679 0.000865586941148724
2.07611300023694 0.000645377672259436
2.09029028841708 0.000513296611652396
2.10446757659722 0.0005421028859769
2.11864486477737 0.000687810477576571
2.13282215295751 0.000741044490798723
2.14699944113765 0.000576011151353161
2.1611767293178 0.000311061650560095
2.17535401749794 0.000127897472208191
2.18953130567809 9.39364174224667e-05
2.20370859385823 0.000189547357050964
2.21788588203837 0.000337468782393985
2.23206317021852 0.000415489176188715
2.24624045839866 0.000407529379159306
2.2604177465788 0.000467796589877717
2.27459503475895 0.000673203268481554
2.28877232293909 0.000859658256448052
2.30294961111923 0.000862078811307472
2.31712689929938 0.000716059714222503
2.33130418747952 0.000508939903037858
2.34548147565967 0.000288365270186371
2.35965876383981 0.000117941947692979
2.37383605201995 3.27343197133302e-05
2.3880133402001 5.99741838190017e-06
2.40219062838024 7.17541629928936e-07
2.41636791656038 5.58312692481189e-08
2.43054520474053 2.8209203572308e-09
2.44472249292067 9.2498968580893e-11
2.45889978110081 1.96798081914361e-12
2.47307706928096 2.71648020386106e-14
2.4872543574611 2.43265244638476e-16
2.50143164564125 1.41748970789777e-18
2.51560893382139 1.13564633667955e-18
2.52978622200153 1.98087326285624e-16
2.54396351018168 2.25211827087402e-14
2.55814079836182 1.6611345189773e-12
2.57231808654196 7.94872531200567e-11
2.58649537472211 2.46756869504345e-09
2.60067266290225 4.96958463719119e-08
2.6148499510824 6.49306884814379e-07
2.62902723926254 5.50375383440824e-06
2.64320452744268 3.02654454998383e-05
2.65738181562283 0.000107972763682523
2.67155910380297 0.000249896654088875
2.68573639198311 0.000375219794851354
2.69991368016326 0.000365502303889133
2.7140909683434 0.000230979572254496
2.72826825652354 9.46969744248695e-05
2.74244554470369 2.51870735869658e-05
2.75662283288383 4.34608679539002e-06
};\label{plot:librispeech-char-based}
\addplot [very thick, palette-2]
table {%
-0.0390492340040308 0.0143538038723533
-0.0205738539880485 0.02142125727255
-0.00209847397206619 0.0241311561233586
0.0163769060439161 0.0225836412382787
0.0348522860598984 0.0182322690503782
0.0533276660758807 0.0167894393042975
0.0718030460918629 0.0164616759683456
0.0902784261078452 0.0161219255955083
0.108753806123828 0.0157742114103423
0.12722918613981 0.0152193796325926
0.145704566155792 0.0148615154656845
0.164179946171774 0.0145399222592108
0.182655326187757 0.0141953801483321
0.201130706203739 0.0136079191696464
0.219606086219721 0.0133042525017712
0.238081466235704 0.0130306397585902
0.256556846251686 0.0127905143857183
0.275032226267668 0.0125231429691813
0.29350760628365 0.0121729203126215
0.311982986299633 0.012194225508638
0.330458366315615 0.0117861006342181
0.348933746331597 0.0115350869846791
0.367409126347579 0.0115451525223066
0.385884506363562 0.0108976055101647
0.404359886379544 0.010290322476282
0.422835266395526 0.0100454937339872
0.441310646411509 0.0104006615480655
0.459786026427491 0.0100024461672047
0.478261406443473 0.00924416995731907
0.496736786459455 0.00908915093353291
0.515212166475438 0.00877402509965138
0.53368754649142 0.00814875951292906
0.552162926507402 0.00789136937598148
0.570638306523385 0.00861723200956459
0.589113686539367 0.00816855567755839
0.607589066555349 0.00820793934769053
0.626064446571332 0.00809920636230281
0.644539826587314 0.00705438763421333
0.663015206603296 0.00698000921483132
0.681490586619278 0.00565383278962989
0.699965966635261 0.00569110450806142
0.718441346651243 0.00631737067885462
0.736916726667225 0.0048358939294873
0.755392106683208 0.00428827081872926
0.77386748669919 0.00349321575028152
0.792342866715172 0.00421919230227779
0.810818246731154 0.00604727218293054
0.829293626747137 0.00413350765917816
0.847769006763119 0.00261411657680491
0.866244386779101 0.00133824407810788
0.884719766795083 0.0023550441301661
0.903195146811066 0.00356313290233603
0.921670526827048 0.00559898675684399
0.94014590684303 0.00566222858471832
0.958621286859013 0.00282999093162825
0.977096666874995 0.0022503592519637
0.995572046890977 0.003050589412602
1.01404742690696 0.00306100128525735
1.03252280692294 0.00419084746350681
1.05099818693892 0.00349076165736073
1.06947356695491 0.000587424232120477
1.08794894697089 0.00114231383654121
1.10642432698687 0.00228734038716405
1.12489970700285 0.000616913179894504
1.14337508701884 2.22733779095575e-05
1.16185046703482 1.88326914530724e-05
1.1803258470508 0.000562966877322223
1.19880122706678 0.00225711797501645
1.21727660708276 0.00120696963975468
1.23575198709875 0.000105402435343825
1.25422736711473 0.000573715878112907
1.27270274713071 0.00226330580192983
1.29117812714669 0.00119247292719797
1.30965350716268 8.37902191313741e-05
1.32812888717866 7.8519483170583e-07
1.34660426719464 9.81297847750177e-10
1.36507964721062 1.63554903137386e-13
1.38355502722661 3.63551212885455e-18
1.40203040724259 1.07772246422049e-23
1.42050578725857 4.2607649930036e-30
1.43898116727455 2.2465057482687e-37
1.45745654729053 1.57967154563784e-45
1.47593192730652 1.48137620486714e-54
1.4944073073225 1.85269206409835e-64
1.51288268733848 3.09284927500393e-69
1.53135806735446 6.1639973186036e-59
1.54983344737045 1.63834691666466e-49
1.56830882738643 5.80749177011816e-41
1.58678420740241 2.74543171160243e-33
1.60525958741839 1.73090007353709e-26
1.62373496743437 1.45536744114177e-20
1.64221034745036 1.63197169167165e-15
1.66068572746634 2.44057352256517e-11
1.67916110748232 4.86755062923197e-08
1.6976364874983 1.29469770122121e-05
1.71611186751429 0.000459268867138411
1.73458724753027 0.00217379738921521
1.75306262754625 0.00147397130542919
1.77153800756223 0.00142237579295108
1.79001338757822 0.00220982472022838
1.8084887675942 0.000497692878287552
1.82696414761018 1.49573760556052e-05
1.84543952762616 5.99501639326508e-08
1.86391490764214 3.20453232069921e-11
1.88239028765813 2.28443205403065e-15
1.90086566767411 2.17185657701264e-20
1.91934104769009 2.75374262597936e-26
1.93781642770607 4.65644909901687e-33
1.95629180772206 1.05008796359193e-40
1.97476718773804 3.15817221911485e-49
1.99324256775402 1.26673346852135e-58
2.01171794777 6.77600923801539e-69
2.03019332778599 4.83394874754239e-80
2.04866870780197 4.59906345068621e-92
2.06714408781795 5.83547306863302e-105
2.08561946783393 9.87465758022945e-119
2.10409484784991 2.22847275159849e-133
2.1225702278659 6.70705579987282e-149
2.14104560788188 2.69212870466935e-165
2.15952098789786 1.44111628469229e-182
2.17799636791384 1.02882507533306e-200
2.19647174792983 9.79542679808448e-220
2.21494712794581 1.24378264919459e-239
2.23342250796179 2.10622582081405e-260
2.25189788797777 1.39140276709868e-265
2.27037326799375 1.34068981320611e-244
2.28884864800974 1.72283253342447e-224
2.30732402802572 2.95254908828786e-205
2.3257994080417 6.74823964931563e-187
2.34427478805768 2.05694734981019e-169
2.36275016807367 8.36171116184649e-153
2.38122554808965 4.53321644599274e-137
2.39970092810563 3.27760835141543e-122
2.41817630812161 3.16043576086879e-108
2.4366516881376 4.06421041892709e-95
2.45512706815358 6.97019150296403e-83
2.47360244816956 1.59423574952768e-71
2.49207782818554 4.86294866250534e-61
2.51055320820152 1.97827259082814e-51
2.52902858821751 1.07327716824453e-42
2.54750396823349 7.76563466918893e-35
2.56597934824947 7.49344161951422e-28
2.58445472826545 9.64328514233057e-22
2.60293010828144 1.65503810379907e-16
2.62140548829742 3.78817726374633e-12
2.6398808683134 1.1563569027921e-08
2.65835624832938 4.70752688688544e-06
2.67683162834537 0.000255583627002292
2.69530700836135 0.00185059999474163
2.71378238837733 0.0017870281432799
2.73225776839331 0.000230138636383846
2.75073314840929 3.95263750088901e-06
2.76920852842528 9.05365092736865e-09
2.78768390844126 2.76578279177047e-12
2.80615928845724 2.83786212789385e-12
2.82463466847322 9.2368434179528e-09
2.84311004848921 4.00970499806375e-06
2.86158542850519 0.000232134970184986
2.88006080852117 0.00179228894536268
2.89853618853715 0.00184550314667182
2.91701156855313 0.0002534316562262
2.93548694856912 4.64137054081912e-06
2.9539623285851 1.13362895197306e-08
2.97243770860108 3.69262260752991e-12
2.99091308861706 1.60412507463006e-16
3.00938846863305 9.29353285361955e-22
3.02786384864903 7.18063344406372e-28
3.04633922866501 7.3991866743184e-35
3.06481460868099 1.01682102809361e-42
3.08328998869698 1.86356421127473e-51
3.10176536871296 4.55494892808423e-61
3.12024074872894 1.48477952523597e-71
3.13871612874492 6.45475460749537e-83
3.1571915087609 3.74228458052358e-95
3.17566688877689 2.89356466255672e-108
3.19414226879287 2.9837943117571e-122
3.21261764880885 4.10340023735817e-137
3.23109302882483 7.52589551687569e-153
3.24956840884082 2.09601255687043e-159
3.2680437888568 2.56322505885188e-143
3.28651916887278 4.18041029144741e-128
3.30499454888876 9.0926481120317e-114
3.32346992890474 2.63755283333046e-100
3.34194530892073 1.02035491177331e-87
3.36042068893671 5.26430034125408e-76
3.37889606895269 3.62217447804586e-65
3.39737144896867 3.32381911718327e-55
3.41584682898466 4.06766055864663e-46
3.43432220900064 6.63882675116918e-38
3.45279758901662 1.44503136613791e-30
3.4712729690326 4.1947155954254e-24
3.48974834904859 1.62392914100291e-18
3.50822372906457 8.38437969480131e-14
3.52669910908055 5.77316683058439e-10
3.54517448909653 5.30147338930789e-07
3.56364986911251 6.49259587867539e-05
3.5821252491285 0.00106042379785753
3.60060062914448 0.00230982919976791
3.61907600916046 0.000670995452263669
3.63755138917644 2.59955250782879e-05
};\label{plot:timit-char-based}
    \addplot [very thick, palette-1,dotted]
table {%
-0.0916935881188112 0.0115423714577064
-0.0747366953600361 0.0148316448686822
-0.057779802601261 0.0174562377577941
-0.0408229098424859 0.0194185174673208
-0.0238660170837108 0.0207230654258067
-0.00690912432493569 0.0213794838689694
0.0100477684338394 0.0214096936043963
0.0270046611926145 0.0208677240627304
0.0439615539513896 0.0198901946483548
0.0609184467101647 0.018780080682703
0.0778753394689398 0.0179499392961506
0.0948322322277148 0.0175340143396034
0.11178912498649 0.0173372270416589
0.128746017745265 0.0171984739879639
0.14570291050404 0.0170667807288954
0.162659803262815 0.016932786221082
0.17961669602159 0.0167962909505146
0.196573588780365 0.0166593653784542
0.213530481539141 0.016523332882107
0.230487374297916 0.0163860633589275
0.247444267056691 0.0162437085895872
0.264401159815466 0.0160970392244186
0.281358052574241 0.0159545994655765
0.298314945333016 0.0158248207643772
0.315271838091791 0.0157045271641528
0.332228730850566 0.0155829003294357
0.349185623609341 0.0154572635554971
0.366142516368116 0.0153365888159793
0.383099409126891 0.0152294431622814
0.400056301885666 0.0151335965127631
0.417013194644442 0.0150366396197509
0.433970087403217 0.014925146762375
0.450926980161992 0.0147947071857443
0.467883872920767 0.0146519571732491
0.484840765679542 0.0145079557116444
0.501797658438317 0.0143716103339946
0.518754551197092 0.0142492464180634
0.535711443955867 0.0141444773933086
0.552668336714642 0.0140513362398209
0.569625229473418 0.0139513081450128
0.586582122232193 0.0138268353470928
0.603539014990968 0.0136767948510008
0.620495907749743 0.0135149315616867
0.637452800508518 0.0133592130681313
0.654409693267293 0.0132232025110467
0.671366586026068 0.0131044720675593
0.688323478784843 0.0129817568600327
0.705280371543618 0.012837666498687
0.722237264302393 0.0126804338405643
0.739194157061168 0.0125296804765937
0.756151049819944 0.0123861316157231
0.773107942578719 0.0122359731396696
0.790064835337494 0.0120820120991444
0.807021728096269 0.0119479689187409
0.823978620855044 0.0118489076706864
0.840935513613819 0.0117737451711548
0.857892406372594 0.0117023098854302
0.874849299131369 0.0116254362382199
0.891806191890144 0.0115426491999191
0.908763084648919 0.0114493993494615
0.925719977407694 0.0113342928069208
0.94267687016647 0.0111916036114803
0.959633762925245 0.0110335844855252
0.97659065568402 0.0108794471872687
0.993547548442795 0.0107313886341366
1.01050444120157 0.010577180771772
1.02746133396034 0.0104176060902378
1.04441822671912 0.0102769335667334
1.0613751194779 0.0101781549472023
1.07833201223667 0.0101118959869095
1.09528890499545 0.010046198924794
1.11224579775422 0.00997045350579852
1.129202690513 0.00990648438736969
1.14615958327177 0.00986064521061723
1.16311647603055 0.00979429183791189
1.18007336878932 0.00966431187655504
1.1970302615481 0.00946786340089589
1.21398715430687 0.00924645338226343
1.23094404706565 0.00906048286702472
1.24790093982442 0.00894214476741467
1.2648578325832 0.00886588690068974
1.28181472534197 0.00878268359936434
1.29877161810075 0.00867200975677688
1.31572851085952 0.00855074893322536
1.3326854036183 0.00845032651090724
1.34964229637707 0.00839073711683584
1.36659918913585 0.00835731393855316
1.38355608189462 0.00830850676172966
1.4005129746534 0.00823131287274471
1.41746986741217 0.00816526306033564
1.43442676017095 0.0081354109111368
1.45138365292972 0.00810074657835641
1.4683405456885 0.00801830050847406
1.48529743844727 0.00791499111438927
1.50225433120605 0.00785691792026154
1.51921122396482 0.00785909635861657
1.5361681167236 0.00786898108193917
1.55312500948237 0.0078275959663405
1.57008190224115 0.0077034930540022
1.58703879499992 0.00750285668467276
1.6039956877587 0.00728249399529375
1.62095258051747 0.00711306002176377
1.63790947327625 0.00699486986404792
1.65486636603502 0.00685596646972215
1.6718232587938 0.00663830738443797
1.68878015155257 0.00635650164601794
1.70573704431135 0.0060998773525589
1.72269393707012 0.00597086515396976
1.7396508298289 0.00599040855181224
1.75660772258767 0.00609633678197494
1.77356461534645 0.00621650745760498
1.79052150810522 0.00629034883882815
1.807478400864 0.00626960426956987
1.82443529362277 0.00615322770706512
1.84139218638155 0.00600450110486986
1.85834907914032 0.00589103487480408
1.8753059718991 0.00580865367912298
1.89226286465787 0.00570585033428959
1.90921975741665 0.00555742072918081
1.92617665017542 0.00516801041830309
1.9431335429342 0.00501491182840534
1.96009043569298 0.00531353786419587
1.97704732845175 0.00570247282400424
1.99400422121053 0.00586582371880227
2.0109611139693 0.00584366279109971
2.02791800672808 0.00572320519068535
2.04487489948685 0.00569271203145325
2.06183179224563 0.00578162101679532
2.0787886850044 0.00581111335073537
2.09574557776318 0.00562952863897905
2.11270247052195 0.00478881366099393
2.12965936328073 0.00384275820880383
2.1466162560395 0.00323419912505138
2.16357314879828 0.00295177738978118
2.18053004155705 0.00280156511643509
2.19748693431583 0.00259941003279172
2.2144438270746 0.00227624680863234
2.23140071983338 0.00193519573827772
2.24835761259215 0.00178949206254851
2.26531450535093 0.00192609455618213
2.2822713981097 0.00214599465889044
2.29922829086848 0.00216291197694552
2.31618518362725 0.00195150330436716
2.33314207638603 0.0017554084369156
2.3500989691448 0.00173190868769759
2.36705586190358 0.00175205376206412
2.38401275466235 0.00160734766112778
2.40096964742113 0.0012630989870939
2.4179265401799 0.000851766861741087
2.43488343293868 0.000533094368486818
2.45184032569745 0.000398969395959353
2.46879721845623 0.000442607306660386
2.485754111215 0.000587923895477302
2.50271100397378 0.00076145290286199
2.51966789673255 0.000931065728821675
2.53662478949133 0.00107921682904094
2.5535816822501 0.00118560098397851
2.57053857500888 0.0012549826165595
2.58749546776765 0.00130737352295862
2.60445236052643 0.00131748657703011
2.6214092532852 0.00121569687694877
2.63836614604398 0.00097011833386838
2.65532303880275 0.000640574279560682
2.67227993156153 0.000338416562831311
2.6892368243203 0.000139408471454113
2.70619371707908 4.41746268540522e-05
2.72315060983785 1.2308081768027e-05
2.74010750259663 1.16087617053348e-05
2.7570643953554 3.84188155955323e-05
2.77402128811418 0.00011310320781159
2.79097818087295 0.00025753182208024
2.80793507363173 0.000462083868133198
2.8248919663905 0.000674869231136969
2.84184885914928 0.000827800294077353
2.85880575190805 0.000862985367484113
2.87576264466683 0.00075155135597407
2.8927195374256 0.000528125658361563
2.90967643018438 0.00029034125853342
2.92663332294316 0.00012679158387462
2.94359021570193 6.27697024670588e-05
2.96054710846071 8.34865919346813e-05
2.97750400121948 0.000168881753308128
2.99446089397826 0.000288364390345525
3.01141778673703 0.000393765933443725
3.02837467949581 0.000442265556597973
3.04533157225458 0.000416472773619026
3.06228846501336 0.000325306337170667
3.07924535777213 0.000204261288602541
3.09620225053091 0.00010141199838014
3.11315914328968 4.72525490276433e-05
3.13011603604846 4.78835584441257e-05
3.14707292880723 0.000100097908202564
3.16402982156601 0.000187027422296664
3.18098671432478 0.000261392337625235
3.19794360708356 0.000268913182821809
3.21490049984233 0.000203379306438654
3.23185739260111 0.000113066226995045
3.24881428535988 4.62047934174682e-05
3.26577117811866 1.38793422650796e-05
3.28272807087743 3.06463273806946e-06
};\label{plot:librispeech-equally-sized}
\addplot [very thick, palette-2,dotted]
table {%
-0.0464653790676541 0.0138802956043254
-0.0274234845544114 0.0202557801543128
-0.00838159004116861 0.0233600644260868
0.0106603044720741 0.0232317631151918
0.0297021989853169 0.0202275369412279
0.0487440934985597 0.0172646851422703
0.0677859880118024 0.0167079524656264
0.0868278825250452 0.0164008027086477
0.105869777038288 0.0160819860227231
0.124911671551531 0.0158360226780451
0.143953566064773 0.0156136702813163
0.162995460578016 0.015362264421536
0.182037355091259 0.015214671032106
0.201079249604502 0.0148418906331574
0.220121144117744 0.0144286494383003
0.239163038630987 0.0142313292423177
0.25820493314423 0.0139149225677751
0.277246827657473 0.013603228633966
0.296288722170715 0.0132470338983509
0.315330616683958 0.0129112139119806
0.334372511197201 0.0126870439896997
0.353414405710444 0.0123794780515279
0.372456300223686 0.0122513863486819
0.391498194736929 0.0120811897786631
0.410540089250172 0.0120336702793793
0.429581983763415 0.0120025612634862
0.448623878276657 0.0116893404357591
0.4676657727899 0.0111655287734362
0.486707667303143 0.0105875245381302
0.505749561816386 0.0101633020441136
0.524791456329629 0.00959592364084638
0.543833350842871 0.00944302550752491
0.562875245356114 0.00977154585405732
0.581917139869357 0.00985126888671024
0.600959034382599 0.00966391240115363
0.620000928895842 0.00938902228500251
0.639042823409085 0.00901261048962589
0.658084717922328 0.00872084075900325
0.67712661243557 0.0080111939052396
0.696168506948813 0.00750910074160065
0.715210401462056 0.00688003301984975
0.734252295975299 0.00713397946918685
0.753294190488542 0.0070573788540771
0.772336085001784 0.00657117057842875
0.791377979515027 0.00580301116207337
0.81041987402827 0.00568933829621653
0.829461768541513 0.00638927321399746
0.848503663054755 0.00639397217436998
0.867545557567998 0.00567266651853844
0.886587452081241 0.00564061262484356
0.905629346594484 0.00584468419215746
0.924671241107726 0.00612331897631334
0.943713135620969 0.00665061328708188
0.962755030134212 0.006454029705453
0.981796924647454 0.00320597846591149
1.0008388191607 0.000392641165752106
1.01988071367394 3.05633078537519e-05
1.03892260818718 0.000442355020225643
1.05796450270043 0.00264885754608005
1.07700639721367 0.00431779507030248
1.09604829172691 0.00313635007755549
1.11509018624015 0.00129371448325147
1.1341320807534 0.000166781894412688
1.15317397526664 5.02548891353324e-06
1.17221586977988 3.3671948641675e-08
1.19125776429312 7.85749333727902e-11
1.21029965880637 2.17620511150751e-08
1.22934155331961 3.63309457120587e-06
1.24838344783285 0.000133790201930703
1.2674253423461 0.00108678618666501
1.28646723685934 0.00194841691300866
1.30550913137258 0.000830871501029803
1.32455102588582 0.000805982528162103
1.34359292039907 0.00212244050079216
1.36263481491231 0.00238618671043168
1.38167670942555 0.00203403798458573
1.4007186039388 0.000629977900084421
1.41976049845204 4.57672731337155e-05
1.43880239296528 7.3745126980288e-07
1.45784428747852 2.6223235513179e-09
1.47688618199177 2.05696625324883e-12
1.49592807650501 5.4594362552696e-15
1.51496997101825 1.90427174561577e-11
1.53401186553149 1.56732066312491e-08
1.55305376004474 2.84549080741225e-06
1.57209565455798 0.00011395329995986
1.59113754907122 0.00100662339030769
1.61017944358447 0.00196145125912235
1.62922133809771 0.000843059866615105
1.64826323261095 7.99299795844666e-05
1.66730512712419 1.67159622008013e-06
1.68634702163744 7.71122633731122e-09
1.70538891615068 7.84667947176549e-12
1.72443081066392 1.76124091638261e-15
1.74347270517716 8.72011097123467e-20
1.76251459969041 9.52347945105581e-25
1.78155649420365 2.29424438844333e-30
1.80059838871689 1.21914145488978e-36
1.81964028323014 1.42902198934047e-43
1.83868217774338 3.69482706364991e-51
1.85772407225662 2.1072677091545e-59
1.87676596676986 2.65103645748981e-68
1.89580786128311 7.35668363313953e-78
1.91484975579635 4.5031752798728e-88
1.93389165030959 6.08031410125763e-99
1.95293354482284 1.81093795075977e-110
1.97197543933608 1.18973850380256e-122
1.99101733384932 1.72412939264547e-135
2.01005922836256 5.51135754140931e-149
2.02910112287581 3.88613343409357e-163
2.04814301738905 5.73188014881306e-169
2.06718491190229 1.48599954863167e-154
2.08622680641553 8.49788102338589e-141
2.10526870092878 1.07194624422923e-127
2.12431059544202 2.98267380020734e-115
2.14335248995526 1.83066510592526e-103
2.16239438446851 2.47846299820852e-92
2.18143627898175 7.40161288838917e-82
2.20047817349499 4.87574166341707e-72
2.21952006800823 7.08476416516545e-63
2.23856196252148 2.27080844835919e-54
2.25760385703472 1.60548405322025e-46
2.27664575154796 2.50381295904975e-39
2.2956876460612 8.61327198535002e-33
2.31472954057445 6.53589146365627e-27
2.33377143508769 1.09398632588267e-21
2.35281332960093 4.03914410154057e-17
2.37185522411418 3.28955531076542e-13
2.39089711862742 5.90956386461994e-10
2.40993901314066 2.34176903784445e-07
2.4289809076539 2.0469304215317e-05
2.44802280216715 0.000394668525625422
2.46706469668039 0.00167854232960545
2.48610659119363 0.00157471692064472
2.50514848570687 0.00032586902572927
2.52419038022012 1.48749002493596e-05
2.54323227473336 1.49773656888023e-07
2.5622741692466 3.32649746735817e-10
2.58131606375985 1.62970632977266e-13
2.60035795827309 1.76117251558903e-17
2.61939985278633 4.19821656724159e-22
2.63844174729957 2.20748707131762e-27
2.65748364181282 2.56036505464049e-33
2.67652553632606 6.55052344245575e-40
2.6955674308393 3.6967536565346e-47
2.71460932535255 4.60188126235826e-55
2.73365121986579 1.26363224131368e-63
2.75269311437903 7.6537871350027e-73
2.77173500889227 1.02259214575749e-82
2.79077690340552 5.42480772346863e-82
2.80981879791876 3.6460670058382e-72
2.828860692432 5.40549330061153e-63
2.84790258694524 1.76773195630761e-54
2.86694448145849 1.27516940535276e-46
2.88598637597173 2.0290359761698e-39
2.90502827048497 7.1216725684184e-33
2.92407016499822 5.51371925362294e-27
2.94311205951146 9.41624364826652e-22
2.9621539540247 3.54716277142409e-17
2.98119584853794 2.94750764207179e-13
3.00023774305119 5.40255446181836e-10
3.01927963756443 2.18430741065322e-07
3.03832153207767 1.94804397588323e-05
3.05736342659091 0.000383225276071012
3.07640532110416 0.00166295281530827
3.0954472156174 0.00159175454810057
3.11448911013064 0.000336079996465585
3.13353100464389 1.56523525131986e-05
3.15257289915713 1.60800342095368e-07
3.17161479367037 3.64388538530962e-10
3.19065668818361 1.82143148678637e-13
3.20969858269686 2.00831283270325e-17
3.2287404772101 4.88450185130112e-22
3.24778237172334 2.62047237389105e-27
3.26682426623658 3.10105368209462e-33
3.28586616074983 8.09486061228669e-40
3.30490805526307 4.66100809582317e-47
3.32394994977631 5.91998670524289e-55
3.34299184428956 1.65856309404597e-63
3.3620337388028 1.02497590233351e-72
3.38107563331604 1.39722293082513e-82
3.40011752782928 6.76288668061931e-83
3.41915942234253 5.19820254957828e-73
3.43820131685577 8.813424271306e-64
3.45724321136901 3.29614682083122e-55
3.47628510588226 2.71918534639687e-47
3.4953270003955 4.94813415529528e-40
3.51436889490874 1.98616128533338e-33
3.53341078942198 1.75856307734054e-27
3.55245268393523 3.43456532099421e-22
3.57149457844847 1.47963979202147e-17
3.59053647296171 1.40608189391945e-13
3.60957836747495 2.94737653224651e-10
3.6286202619882 1.36279652779132e-07
3.64766215650144 1.38994241307613e-05
3.66670405101468 0.000312703676989768
3.68574594552793 0.00155181362439366
3.70478784004117 0.0016986993663397
3.72382973455441 0.000410169988783333
3.74287162906765 2.18464879057956e-05
};\label{plot:timit-equally-sized}

\coordinate (legend) at (axis description cs: 0.5, 1.05);
\end{axis}

\matrix [
    draw,
    matrix of nodes,
    anchor=south,
    column sep=5pt,
    row sep=2pt,
    nodes={inner sep=0},
    column 1/.style={anchor=base west}
] at (legend) {
                     & Timit & LibriSpeech \\
    (2) equally sized & \ref{plot:timit-equally-sized}   & \ref{plot:librispeech-equally-sized}   \\
    (3) character-based & \ref{plot:timit-char-based}   & \ref{plot:librispeech-char-based}   \\
};
\end{tikzpicture}

%% file: tikz/tcpwer_over_collar.tex
\begin{tikzpicture}[
    label/.style={fill=white,inner sep=0,inner xsep=2px,sloped}
]
\begin{groupplot}[group style={group size= 1 by 2, x descriptions at=edge bottom, vertical sep=3mm}]
\nextgroupplot[
    height=5cm,
    width=8cm,
    log basis x={10},
    ylabel=Error Rate in \si{\percent},
    xmajorgrids,
    xmin=0.5, xmax=100,
    xminorgrids,
    xmode=log,
    ymajorgrids,
    ymin=0.1, ymax=0.12,
    yminorgrids,
    ymode=log,
    ytick={0.1,0.11,0.12,0.13,0.14,0.15,0.2},
    yticklabels={10,11, 12, 13, 14, 15,20}
    ]
    \path[name path=d1] (rel axis cs:0,.97) -- (rel axis cs:1,.97);
    \path[name path=d2] (rel axis cs:0,.5) -- (rel axis cs:1,.5);
    
    \addplot [very thick, palette-1, name path=p0] table[x=hyp_collar,y=0.0,col sep=comma] {tikz/tcpwer_over_collar.csv}; 
    \path [name intersections={of=p0 and d1,by={E1}},name intersections={of=p0 and d2,by={E2}}] (E1) -- (E2) node[label,pos=0,anchor=west,text=palette-1]{\footnotesize TS-SEP};
    \addplot [very thick, palette-2, name path=p1] table[x=hyp_collar,y=5.0,col sep=comma] {tikz/tcpwer_over_collar.csv};
    \path [name intersections={of=p1 and d1,by={E1}},name intersections={of=p1 and d2,by={E2}}] (E1) -- (E2) node[label,pos=0,anchor=west,text=palette-2]{\footnotesize + \SI{5}{s} pauses};
    \addplot [very thick, palette-3, name path=p2] table[x=hyp_collar,y=10.0,col sep=comma] {tikz/tcpwer_over_collar.csv}; 
    \path [name intersections={of=p2 and d1,by={E1}},name intersections={of=p2 and d2,by={E2}}] (E1) -- (E2) node[label,pos=0,anchor=west,text=palette-3]{\footnotesize + \SI{10}{s} pauses};
    \addplot [very thick, palette-4,name path=p3] table[x=hyp_collar,y=20.0,col sep=comma] {tikz/tcpwer_over_collar.csv}; 
    \path [name intersections={of=p3 and d1,by={E1}},name intersections={of=p3 and d2,by={E2}}] (E1) -- (E2) node[label,pos=0,anchor=west,text=palette-4]{\footnotesize + \SI{20}{s} pauses};
    \addplot [very thick, palette-5, name path=p4] table[x=hyp_collar,y=40.0,col sep=comma] {tikz/tcpwer_over_collar.csv}; 
    \path [name intersections={of=p4 and d1,by={E1}},name intersections={of=p4 and d2,by={E2}}] (E1) -- (E2) node[label,pos=0,anchor=west,text=palette-5]{\footnotesize + \SI{40}{s} pauses};
    
    \addplot [very thick, palette-6, dotted]
    table {%
    0.0707945784384138 0.100994
    141.253754462275 0.100994
    }node[pos=0.25,anchor=south west,yshift=-1mm]{\footnotesize cpWER (\SI{10.1}{\percent})};
    \addplot [very thick, palette-7, dotted]
    table{
    0.0707945784384138 0.1053
    141.253754462275 0.1053
    } node[pos=0.25,anchor=south west,yshift=-1mm]{\footnotesize desired WER (\SI{10.5}{\percent})};
    
\nextgroupplot[
    height=4cm,
    width=8cm,
    height=4cm,
    log basis x={10},
    xlabel=collar size $c$ in \si{\second},
    ylabel style={align=center},ylabel={\%matchings disallowed \\ by collar $\collar$},
    xmajorgrids,
    xminorgrids,
    xmin=0.5, xmax=100,
    xmode=log,
    xtick={1,3,5,10,20,40,100},
    xticklabels={1,3,5,10,20,40,100},
    ymajorgrids,
    ymin=0, ymax=2,
]
\addplot [very thick, black] table[x=hyp_collar,y=disallowed_matchings,col sep=comma] {tikz/tcpwer_over_collar.csv};
\end{groupplot}
    \end{tikzpicture}

%% file: tikz/profiling.tex
\begin{tikzpicture}
    \begin{axis}[
        width  = 0.5*\textwidth,
        height = 0.2*\textwidth,
        major x tick style = transparent,
        ybar=2*\pgflinewidth,
        bar width=14pt,
        ymajorgrids = true,
        ylabel = {Exec. Time in \si{\second}},
        xtick = data,
        xticklabels={{Libri-CSS\\(60$\times$\SI{10}{\minute})},CHiME-6\\(2$\times$\SI{134}{\minute}), Mixer 6\\(59$\times$\SI{15}{\minute}),DiPCo\\(5$\times$\SI{33}{\minute})},
        xticklabel style={align=center},
        scaled y ticks = false,
        enlarge x limits=0.25,
        ymin=0,
        legend cell align=left,
        xmin=0.2,
        xmax=2.8,
    ]
\addplot[style={palette-1,fill=palette-1,mark=none}]
    coordinates {(0, 0.9890878466800002) (1,7.527012406040001) (2,2.4083869142) (3,0.9663239494600002)};

\addplot[style={palette-2,fill=palette-2,mark=none}]
    coordinates {(0, 0.6675641785199999) (1,1.0463411656600001) (2,1.0033501559400002) (3,0.38833911678)};

        \legend{cpWER, tcpWER}
    \end{axis}
\end{tikzpicture}